\documentclass[letterpaper, 10 pt, conference]{ieeeconf}

\IEEEoverridecommandlockouts                              

\usepackage{graphicx} 
\usepackage{epsfig} 
\usepackage{mathptmx} 
\usepackage{times} 
\usepackage{amsmath} 
\usepackage{amssymb}  
\usepackage{xcolor}
\usepackage{algorithmic}
\usepackage{algorithm}
\usepackage{cite}
\usepackage{bm}
\usepackage[export]{adjustbox}
\usepackage{subcaption}

\DeclareUnicodeCharacter{2212}{-}
\DeclareUnicodeCharacter{221E}{$\infty$}
\DeclareUnicodeCharacter{03BC}{$\mu$}

\title{\LARGE \bf
Robust Control of a Multi-Axis Shape Memory Alloy-Driven Soft Manipulator}

\author{Zach J. Patterson$^{1}$, Andrew P. Sabelhaus$^{1}$, and Carmel Majidi$^{1}$

\thanks{*This work was supported by the Office of Naval Research (ONR) under grant $\#$: N00014-17-1-2063 (Program Manager: Dr. Tom McKenna), NOPP grant $\#$: N00014-18-12843 (Program Manager: Reginald Beach), and an Intelligence Community Postdoctoral Research Fellowship through the Oak Ridge Institute for Sciece and Education (Corresponding author: Zach J. Patterson.) }

\thanks{$^{1}$Zach J. Patterson, Andrew P. Sabelhaus and Carmel Majidi are with the Soft Machines Lab, Department of Mechanical Engineering, Carnegie Mellon University, Pittsburgh, PA 15213 USA}
}

\begin{document}

\maketitle
\thispagestyle{empty}
\pagestyle{empty}

\begin{abstract}

Control of soft robotic manipulators remains a challenge for designs with advanced capabilities and novel actuation.
Two significant limitations are multi-axis, three-dimensional motion of soft bodies alongside actuator dynamics and constraints, both of which are present in shape-memory-alloy (SMA)-powered soft robots.
This article addresses both concerns with a robust feedback control scheme, demonstrating state tracking control for a soft robot manipulator of this type.
Our controller uses a static beam bending model to approximate the soft limb as an LTI system, alongside a singular-value-decomposition compensator approach to decouple the multi-axial motion and an anti-windup element for the actuator saturation.
We prove stability and verify robustness of our controller, with robustness intended to account for the unmodeled dynamics.
Our implementation is verified in hardware tests of a soft SMA-powered limb, showing low tracking error, with promising results for future multi-limbed robots.

\end{abstract}

\section{INTRODUCTION}\label{introduction}



While the deployment and use of soft robots are promising because of the dynamical stability and safety inherent in their flexible structure (a concept often referred to a morphological intelligence) \cite{laschi_soft_2016}, they are substantially more challenging to control compared to traditional robots \cite{rus_design_2015}. This issue stems from the inherent difficulty introduced in modeling a continuously deforming system, as well as the often complex dynamics of actuators typically used in soft robots \cite{trivedi_soft_2008,rich_untethered_2018}. 
For multi-axis soft robots that move in 3D, these modeling and feedback challenges are even more significant.
State-of-the-art soft robot control, with state feedback, is limited to planar motions with simple force-input actuator models \cite{della_santina_model-based_2020}.


This article contributes a novel robust feedback control architecture for multi-axial soft robot limbs actuated with shape-memory-alloy (SMA) wires.
We simultaneously address the two challenges of actuator dynamics with saturation in addition to three-dimensional motion, and provide a stability proof and robustness verification for our controller.
Our approach is verified in hardware, and demonstrates low-error trajectory tracking.

Prior work has addressed similar problems, but each with assumptions that do not apply to our goal.
Most prior work in soft robot control has been limited to pneumatic/hydraulic actuators \cite{george_thuruthel_control_2018}, and none have combined state feedback models with three-dimensional motion in hardware.
With pneumatic actuation, model-based state feedback with verified stability has been performed for single-axis bending \cite{della_santina_model-based_2020,cao_model-based_2021}.
Approaches without stability verification have been used in three dimensional motion, including learned models \cite{hyatt_model-based_2019,bruder_data-driven_2021,thuruthel_model-based_2019} and open-loop motions \cite{bern_soft_2020}.
With smart and thermal actuators, controllers for the planar case have been based on system identification \cite{doroudchi_tracking_2021}, though these do provide some evidence that internal actuator states may be neglected in closed-loop.
Finally, some past work has shown three-dimensional feedback for SMA-powered soft manipulators \cite{yang_design_2019}, but these only control per-actuator and not in state or task space of the body.

\begin{figure*}[th]
\centering
\includegraphics[width=1.0\textwidth]{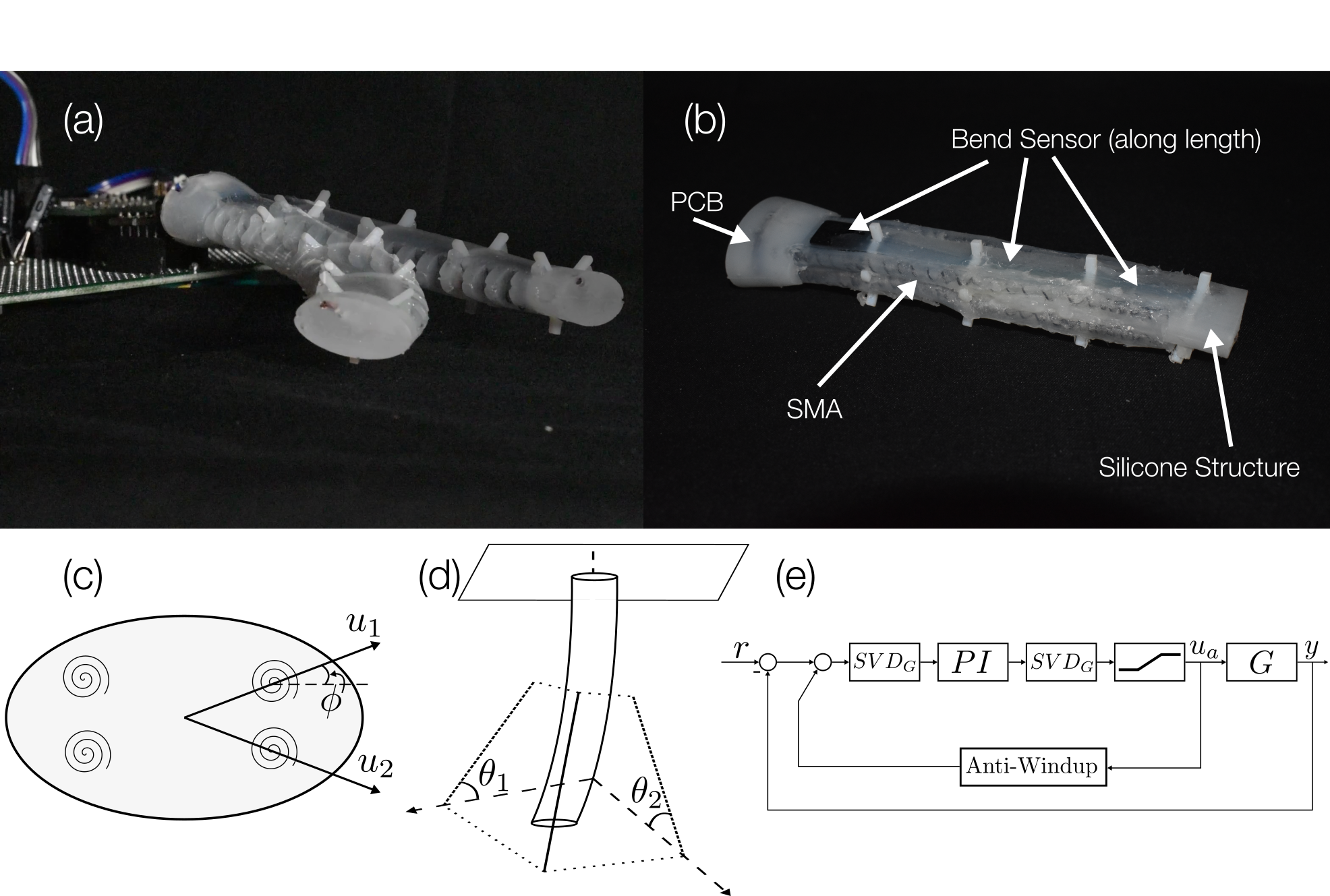}
\caption{(a) Image of a robot manipulator tracking a reference bend angle. (b) Labelled image of critical components of the soft robotic manipulator. (c) Schematic of the input coordinates, $u_i$, which are along the axes of diagonal SMA pairs. (d) Schematic of the the output coordinates, $\theta_i$, which are bend angles in the yaw and pitch directions. (e) Simplified block diagram of control structure. }
\label{fig:fig1}
\end{figure*}

We employ a model-based method to control tip position of a soft manipulator that is actuated using shape memory alloy (Fig. \ref{fig:fig1}a). 
Our particular choice of mechanics approximation is a statics model, not dynamics, which gives a linear time-invariant (LTI) transfer function for the body pose while \textit{not} imposing the requirement of constant-curvature segments.
The tools of robust control theory can be used account for model mismatch in place of a dynamics model \cite{wang_robust_1992}, and in this form, can address the two more salient phenomena of actuator saturation \cite{tarbouriech_stability_2011} and MIMO state feedback \cite{skogestad_multivariable_2007}.
This approach does not require the computational complexity of e.g. model-predictive control in soft and flexible robots \cite{hyatt_model-based_2019,sabelhaus_model-predictive_2021}.
Similarly, we employ an extremely simple model of applied force as our pulse-width-modulated (PWM) control signal to our SMA wires.
The constitutive models of these and other smart and thermal actuators typically require internal states and dynamics \cite{huang_shape_2020} and have therefore received less attention in practical modeling and feedback \cite{rich_untethered_2018}.
We hypothesize that an extremely simple input model, combined with feedback, will be effective for control of our SMAs as well as other thermal actuators.

The contribution of the article is a control framework that enables multi-axis 3D motion of a soft robot limb with shape memory alloy actuators, composed of the following:
\begin{enumerate}
    \item A singular value decomposition (SVD) control method for MIMO motions of a soft robotic manipulator,
    \item A stability proof and robustness analysis of this manipulator with input constraints,
    \item Validation of the controller on an SMA-powered soft robot limb.
\end{enumerate}


The article proceeds as follows. In Sec. \ref{system} we present the design of the manipulator that we seek to control and derive our LTI transfer function. In Sec. \ref{control_design}, we present the design of the controller. In Sec. \ref{proof}, a proof of robust stability is presented. In Sec. \ref{results}, we present results of testing the controller on hardware -- including successful trajectory tracking -- and we compare to other control options. Finally, in Sec. \ref{discussion}, we contextualize the work in the broader context of soft robotic control and discuss areas of improvement that can potentially be addressed in future work.

\section{Manipulator Design}\label{system}

Referring to Fig. \ref{fig:fig1}b, the soft robot manipulator is composed of a silicone elastomer appendage embedded with two antagonistic pairs of SMA coils and a dielectric elastomer bend sensor.  The actuator design is based on a soft robot limb design previously introduced by the authors for an untethered brittlestar-inspired robot  \cite{patterson_untethered_2020}. The four SMA springs are arranged in a rectangular configuration and the silicone rubber has an elliptical cross section (Fig. \ref{fig:fig1}c). A soft capacitive 2-axis angular displacement sensor (Bendlabs) is embedded in the center of the manipulator. A custom PCB containing a connector and MOSFETs for driving the SMAs is embedded in the proximal end. 

The SMA-powered appendage is manufactured as follows. First, the electronics are soldered to the PCB. Next, the four SMA springs are attached to the PCB. Attaching SMA to any soft material can be particularly challenging and in this case we feed the end of the SMA through a via in the PCB, crimp the end, and solder the crimp to the board, providing a robust mechanical and electrical connection. The bend sensor is then soldered to the PCB. This assembly is fed through a series of 3D printed "ossicles" (inspired by the vertebrate-like structures in brittle star limbs) \cite{tomholt_structural_2020} that hold the sensor and SMA coils in place. At this stage, optional silicone food tubing (3mm ID and 4mm OD) can be threaded onto the SMA and sealed with SilPoxy. This tubing prevents silicone from fully encasing the SMA, causing a substantial reduction in the electrical power input needed to induce a given tip displacement. Finally, the distal ends of the SMA are crimped together with a power wire. The completed assembly is then placed in a 3D printed injection mold and DragonSkin 10 Slow is injected. 

The limb is connected to a perforated circuitboard that contains a Laird BL652 SoC which has a Bluetooth enabled nrF52832 microcontroller on board. This controller receives commands via Bluetooth from a nrF52 development kit connected to a computer. The controller drives the SMA MOSFET gates with a pulse width modulated voltage signal. In the controller, these PWM signals will ultimately serve as the inputs of the plant and the outputs are the yaw and pitch bend angles measured by the bend sensor. 

We will now discuss the system representation used to derive a LTI MIMO transfer matrix. Rather than considering each SMA coil as an input, we decided to consider diagonal SMAs to be the same input with one positive and one negative. Fig. \ref{fig:fig1}c shows the inputs of the plant and Fig. \ref{fig:fig1}d shows the outputs of the plant, including their directions. Next, the bending structure is modeled. We use a simplified static model in which the manipulator is a cantilevered Euler Bernoulli beam subject to a constant moment, $M = F*d$, where d is the distance from the SMA to the center of the beam and $F$ is the actuator force. Here, we simply consider $F$ and our commanded PWM signal, called $u$, to be one and the same. The equation for the bend angle, $\theta$, in response to $u$ is then

\begin{equation}
    \theta = \frac{FdL}{EI}\,,
\end{equation}
where $L$ is the length of the manipulator, $E$ is the Young's Modulus, and $I$ is the area moment of inertia. While Young's Modulus doesn't capture the constitutive response of elastomers under large strain, it's adequate for approximating the stress-strain relationship within a linearized small strain regime \cite{larson_can_2016}. We estimate the Young's Modulus of DragonSkin 10 to be $0.19 MPA$ \cite{ainla_soft_2017} and we estimate the Young's Modulus of the 35 ShA rubber tubing to be $1.4 MPA$ (using the Gent Model \cite{gent_relation_1958}). The Young's Modulus used in our model is then approximated as the average of these two values. The cross section of the actuator is approximated as a rectangle with dimensions 16.4$\times$8 mm for the purposes of calculating $I$. These bending relationships are then calculated and arranged in the static gain transfer function matrix:

\begin{equation}
    G(s) = 
    \begin{pmatrix}
        \frac{L d_x cos(\phi)}{EI_y} & \frac{L d_x cos(\phi)}{EI_y}\\
        \frac{L d_y sin(\phi)}{EI_x} & -\frac{L d_y sin(\phi)}{EI_x}
    \end{pmatrix}
\end{equation}  
where row 1 corresponds to pitch and row 2 corresponds to yaw, $d_x$ and $d_y$ are the moment arms of the SMAs for each axis, and $I_x$ and $I_y$ are the area moments of inertia for bending along each axis. 

\section{Control System}\label{control_design}
We will now discuss the design of the control system. The design is based on concepts from robust MIMO control system design. Overall, the system has three critical elements. See Fig. \ref{fig:fig1}e for a high level block diagram. The first important element is the Singular Value Decomposition (SVD) controllers, which is a pre- and post-compensator design meant to shape the plant. The second is a diagonal PI controller. The third is an anti windup feedback loop known as a Hanus conditioner. The design and rationale for each of these sub-units will be presented in detail, along with the final design for the controller.
\subsection{SVD Controller}
When working with MIMO systems, a basic approach to dealing with interactions is to apply \textit{compensators} to reshape the plant $G$ and make interaction easier to deal with \cite{skogestad_multivariable_2007}. An example of a pre-compensator, $W(s)$, is 
\begin{equation}\label{precomp}
    G_s(s) = G(s)W(s).
\end{equation}
The corresponding controller for this shaped plant is then 
\begin{equation}
    K(s) = W(s)K_s(s)
\end{equation}
An SVD is a special class of compensator design in which the plant is shaped by pre- and post-compensators from the singular value decomposition of the plant at a frequency of interest. For example, for the plant at frequency $\omega$, we can make the approximation $G(j\omega)=U\Sigma V$, where $U$ and $V$ are matrices specifying the singular vectors and $\Sigma$ is a diagonal matrix of the singular values. The plant is then shaped to $UGV$ and the corresponding controller is given by:
\begin{equation}
    K(s) = VK_s U
\end{equation}
where $K_s$ is a diagonal controller. A good choice of $K_s$, according to \cite{skogestad_multivariable_2007} is:
\begin{equation}
    K_s = l(s)\Sigma^{-1} \,,
\end{equation}
where $l(s)$ is some chosen controller. This results in a decoupling design, which means that the controller is controlling the plant along the axes of its singular values instead of along some other, usually naive, choice of axes. Hovd et al. found that this structure is optimal for plants consisting of symmetrically interconnected subsystems, which is similar to the control problem here \cite{hovd_svd_1997}.

\subsection{PI Controller}
\begin{figure}[t]
\centering
\includegraphics[width=0.45\textwidth]{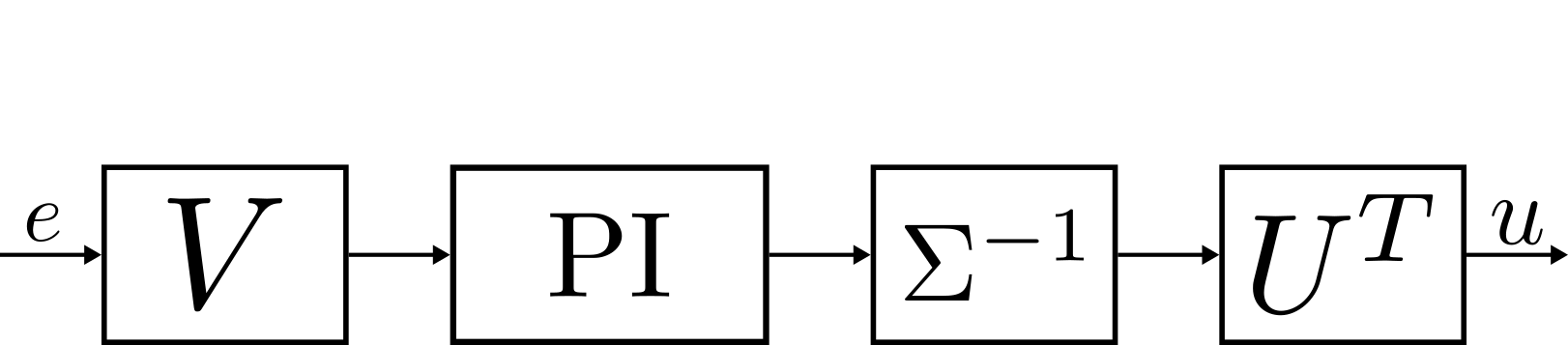}
\caption{Nominal controller for our system including a PI block and SVD pre- and post-compensators.}
\label{fig:ctrlblk}
\end{figure}

The SVD controller design inherently allows flexibility when specifying the controllers at its heart (called $l(s)$ above). For this work, we choose PI controllers as a reasonable first choice thanks to its broad effectiveness, ease of use, and familiarity. Specifically, we choose the PI controller of the following form:
\begin{equation}
    l(s) = K_P(1 + \frac{K_I}{s})\,
\end{equation}
where $K_P$ is the proportional gain and $K_I$ is the integral gain. In principle, we could choose a different PI controller for each of our two singular value directions, but in this case the controllers are chosen to have the same gains because of the symmetry of the problem. At this point, our nominal controller has the form shown in Fig. \ref{fig:ctrlblk}.

\subsection{Anti-Windup}
For systems such as the one we seek to control, the above control scheme would fail on its own. This is due to actuator saturation, in which the controller commands a value of the system input, $u_c$, that is beyond the limits of the real actuator. 
\begin{equation}\label{nonlin}
    u_c = sat(u)
\end{equation}
where
\begin{equation} \label{eq1}
u_{c,i} = \begin{cases}
 u_i & |u| \leq 1 \\
 sign(u_i) & |u_i| > 1.
\end{cases}\notag
\end{equation}

In response, any integral terms in the controller accumulate error. Eventually, and in some cases very rapidly, this accumulated error will come to dominate the control system. The problem is easily dealt with in a classical single-input single-output (SISO) arrangement by adding an extra term into the integrator to eliminate the extra error being accumulated due to saturation \cite{astrom_feedback_2010}. However,  MIMO windup is slightly more complicated to deal with because of interactions in the plant and because of directionality. That said, there are a variety of approaches to dealing with these issues in the literature \cite{tarbouriech_anti-windup_2009}.

A solution to the windup problem for a general MIMO system was developed by Hanus et al \cite{hanus_conditioning_1987}. A nominal controller is specified by the state space form:

\begin{equation}
    \dot{v} = Av + Be
\end{equation}
\begin{equation}
    u_c = Cs + De.
\end{equation}

This controller can be modified to account for windup by augmenting it with a new term, $H$:

\begin{equation}\label{awbt1}
    \dot{v} = Av + Be + H(u_a - u_c)
\end{equation}
\begin{equation}\label{awbt2}
    u_c = Cs + De
\end{equation}
where $u_a$ is the actual (potentially limited) signal applied to the plant. Notice that we can substitute (\ref{awbt2}) into (\ref{awbt1}) to yield

\begin{equation}
    \dot{v} = (A - HC)v + (B - HD)e + Hu_a
\end{equation}
\begin{equation}
    u_c = Cs + De.
\end{equation}
Hanus showed that by choosing $H$ to eliminate the effect of $e$ on the controller states, we can eliminate windup \cite{hanus_conditioning_1987}. Such a method corresponds to the following choice of H:
\begin{equation}
    H = BD^{-1}.
\end{equation}

This approach, usually referred to as Hanus conditioning, successfully eliminates the windup, but there is a final subtlety to address. When the inputs are not saturated, controller performance is nominal, but when an input saturates, the direction of the control changes from the intent of the controller \cite{braatz_stability_1993}. This can be a big deal for certain systems (i.e. with a high condition number). To deal with this problem, we simply adjust the input vector $u_c$ by its largest element:

\begin{equation} \label{eq1}
u_c = Nu = \begin{cases}
 u & \|u\|_{\infty} \leq 1 \\
 \frac{u}{\|u\|_{\infty}} & > 1
\end{cases}.
\end{equation}

\subsection{Overall Controller Structure}
\begin{figure}[t]
\centering
\includegraphics[width=1.0\columnwidth]{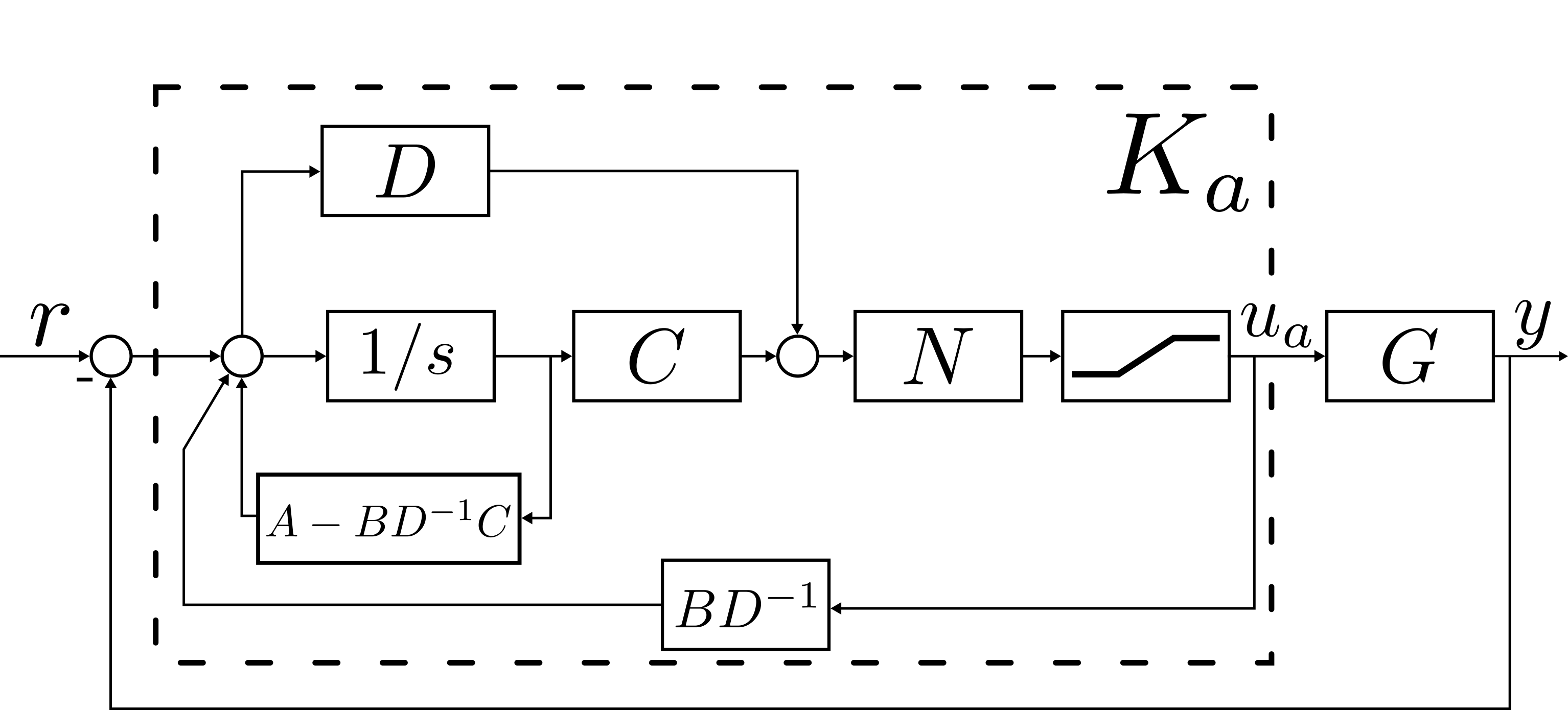}
\caption{Block Diagram of the full closed loop system. $A, B, C, D$ refer to the state space matrices of the nominal controller, $BD^{-1}$ is the anti-windup block, $N$ is the nonlinear adjustment to maintain the direction of the commanded input, and the block directly before the plant, $G$, is the nonlinear operator that constrains the inputs delivered to the plant to [-1,1].}
\label{fig:blk1}
\end{figure}

The procedure for incorporating the above elements is as follows. First, the SVD is performed on the static gain matrix of the plant. The matrices from the SVD are used to create the following controller: 
\begin{equation}
    K(s) = V\begin{pmatrix}
         K_P(1 + \frac{K_I}{s}) & 0\\
        0 &  K_P(1 + \frac{K_I}{s})
    \end{pmatrix}\Sigma^{-1}U\,.
\end{equation}
This nominal controller is then put into state space form for incorporation of the Hanus conditioner exactly according to the equations in the previous section. See the block diagram in Fig. \ref{fig:blk1} for a full specification of the design. Note that the transfer functions $A, B, C, D$ refer to the state space matrices of the nominal controller above.

\section{Proof of Stability}\label{proof}
Ordinarily for analyzing the closed loop stability of the system specified above, we would apply the tools of robust control theory directly and straightforwardly (e.g. with MATLAB's Robust Control Toolbox) to determine if the controller is stable. Unfortunately, it is not immediately obvious how this can be done with the saturation nonlinearity, which is not readily represented in that framework. Here, we will follow an approach proved by Campo and Morari \cite{campo_robust_1990} to indicate robust stability given such a nonlinearity. First we will lay out some preliminary definitions and terminology before demonstrating that our chosen closed loop control system is robustly stable for some set of controller gains. 

First, we note that the nonlinearity described by (\ref{nonlin}) can be bounded by a cone (see \cite{campo_robust_1990} for details. Now, we define a general feedback interconnection shown in Fig. \ref{fig:interconnection}. The matrix $M(s)$ is a LTI transfer function matrix and $\Delta$ is a block diagonal operator representing the uncertainty of the system. It can include cone-bounded nonlinear blocks as well as linear blocks that can be used to model other uncertainties in the model, such as uncertain parameters or uncertain plant dynamics. Therefore, our $\Delta$ is an operator from the set $\bm{\Delta}$:

\begin{figure}[t]
\centering
\includegraphics[width=0.45\textwidth]{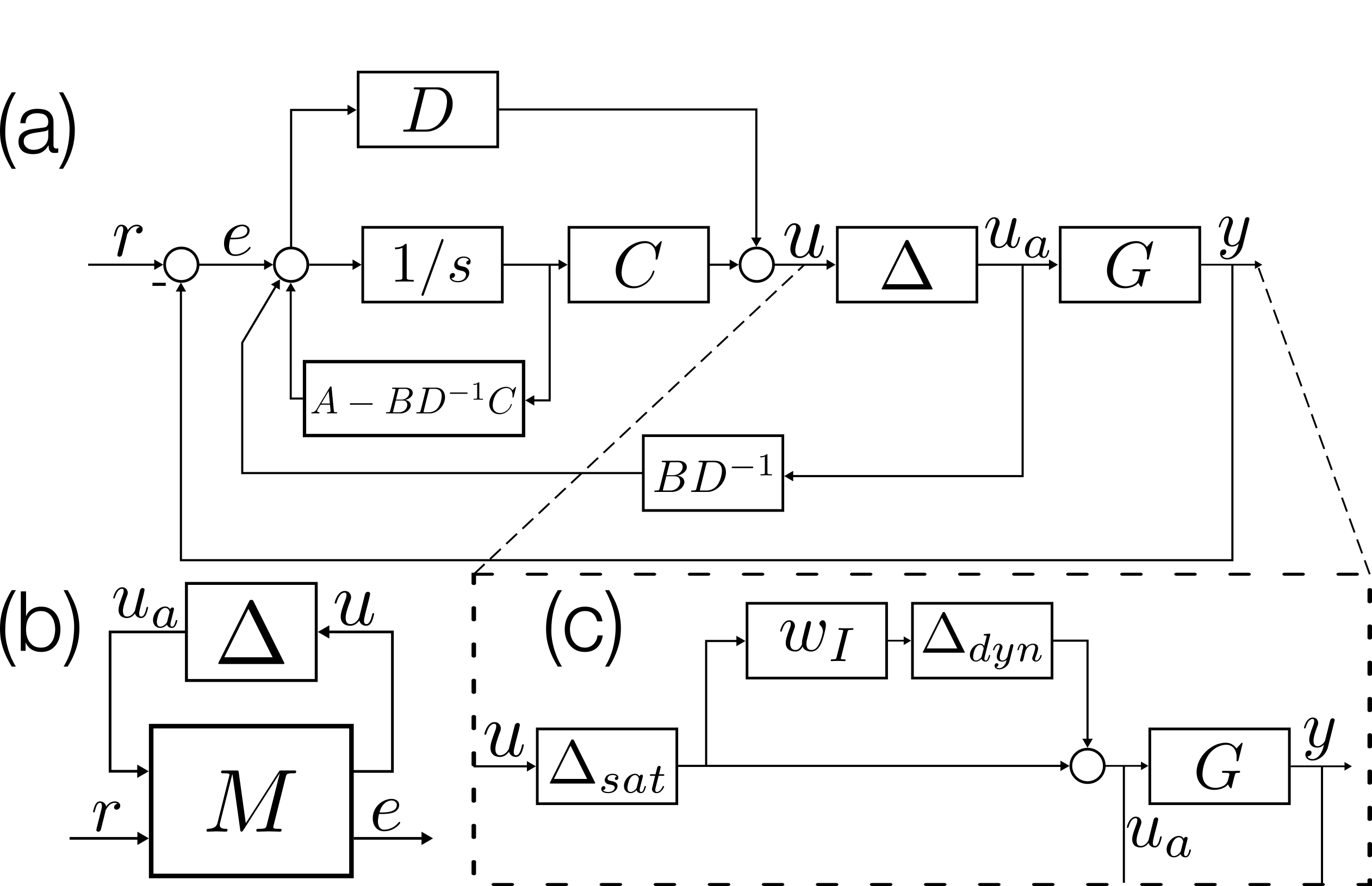}
\caption{(a) Modified block diagram, replacing the nonlinearities with a cone-bounded uncertainty, $\Delta$. (b) Standard $M\Delta$ interconnection for stability analysis. The block diagram in (a) is put into this form for our analysis. (c) Inclusion of an additional multiplicative $\Delta$ block for unmodeled dynamics.}
\label{fig:interconnection}
\end{figure}

\begin{equation}
    \bm{\Delta} = \{\Delta|\Delta = diag(\Delta_1,...,\Delta_n, \Delta_{n+1},...,\Delta_m)\}
\end{equation}

\noindent where $\Delta_1,...,\Delta_n$ are cone-bounded nonlinear operators and $\Delta_{n+1},...,\Delta_m)$ are LTI operators with $\bar{\sigma}(\Delta_i) \leq 1 \forall i = n+1,...,m$.

We can bring our block diagram into this form by "pulling out" the nonlinearity (now represented by $\Delta$), and deriving our transfer function $M(s)$ from the resulting diagram (Fig. \ref{fig:interconnection}b). Once we have that information, following Campo and Morari, we can prove stability:

\textit{Theorem 1}: The system in Fig. \ref{fig:interconnection}b is stable $\forall$ $\Delta \in \bm{\Delta}$ if:

\begin{enumerate}
\item $\begin{aligned}[t]
    M(s) \mbox{ is stable.}
\end{aligned}$
\item $\begin{aligned}[t]
    \exists \beta < 1 \mbox{ s.t. }inf_{W \in \mathcal{W'}}\|WM_{11}(s)W^{-1}\|_{\infty} \leq \beta,
\end{aligned}$
\end{enumerate}

where 

\begin{equation}\label{simp}
    \mathcal{W'} \equiv \{W|W \in \mathcal{W} \mbox{ and } W \in \mathcal{C}^{nxn}\},
\end{equation}
\begin{equation}\label{fullF}
    \mathcal{W} \equiv \{W|W\Delta W^{-1}\in \bm{\Delta}\quad \forall \Delta \in \bm{\Delta}\}.
\end{equation}
Eq. (\ref{simp}) represents a reduction of Eq. (\ref{fullF}) proposed by Campo to make the problem tractable. It relies upon the well known fact from robust control theory that the structure of $W$ must be compatible with the structure of $\Delta$. For example, if $\Delta$ is a scalar times identity block, $W$ must be full; if $\Delta$ is diagonal, $W$ must be diagonal; and if $\Delta$ is full, $W$ must be scalar times identity. 

We now have all pieces necessary to solve the above problem except for the structure to choose for our cone bounded operators $\Delta_1,...,\Delta_n$. Because our nonlinearity is two dimensional (a saturation for each input), we know that the structure must be diagonal, and since we only have two inputs, the case of block diagonal and scalar times identity are equivalent. Therefore, our specific element of $\bm{\Delta}$, $\Delta_{sat}$, has a scalar times identity structure corresponding to a full $W$.

We now have everything necessary to perform the calculations. The above problem forms a Linear Matrix Equality (LMI - see \cite{boyd_linear_1994}), which is a convex optimization problem that is readily solved with computational tools. Here, we use MATLAB to perform the solution. The above LMI is equivalent to solving:

\begin{gather}\label{lmi}
        G(s) = 
    \begin{bmatrix}
        A^T Q + QA & QB-C^T W \\
        B^T Q - WC & \delta I - 2W - WD - D^T W
    \end{bmatrix} \leq 0, \notag\\
    Q > 0, \\ 
    \delta > 0,\notag
\end{gather}  
where $A$, $B$, $C$, and $D$ are the state space matrices of $M_{11}(s)$ and $Q$ is a symmetric matrix. The procedure is then to check $M(s)$ for stability, solve \ref{lmi} for $W$, $Q$, and $\delta$, plug $W$ into Condition 2 of Theorem 1, and check that the resulting value of $\beta$ is less than one.

For our class of problems, the system's closed loop properties and thus $\beta$ depends on our choice of PI gains. In general, increasing $K_P$ will tend to bring the system closer to instability. Using the calculations above, we found that the largest gain for which the system is robustly stable ($\beta < 1$) is $K_P = 2.0$. Although $\beta$ is an upper bound of the structured singular decomposition $\mu$ \cite{skogestad_multivariable_2007}, for this analysis it is conservative as the memoryless nature of the nonlinearity is not accounted for \cite{kothare_multiplier_1999}. 

Next, we incorporate an additional uncertainty into the plant. In this case, we incorporate unmodelled dynamics as a multiplicative uncertainty with a weight of the form

\begin{equation}
    w_{dyn}(s) = \frac{\tau s + r_0}{(\tau/r_{\infty})s + 1}\,,
\end{equation}
where $r_0$ is the relative uncertainty at steady state, $1/\tau$ is the frequency at which the relative uncertainty reaches 100\%, and $r_{\infty}$ is the high frequency magnitude of the weight.  Here, we choose $r_0 = 0.1$, $r_{\infty} = 1.5$, and $\tau = 0.1$
We can now add the new uncertainty weights, along with a new $\Delta_{dyn}$ block, to the block diagram in front of $G$ as shown in Fig. \ref{fig:interconnection}c. We can then re-derive $M(s)$ based on our new expression for $\Delta$, which now includes an additional 2x2 diagonal block representing $\Delta_{dyn}$ as well as the original $\Delta$ representing the saturation nonlinearity, which is now referred to as $\Delta_{sat}$. For this case, using the same gains as before, we get $\beta = 4.7793$.
Including the unmodelled dynamics therefore unsurprisingly causes a large increase in $\beta$. We find that the largest gain that still passes the test is $K_P = 0.5$

\begin{figure}[t]
\centering
\includegraphics[width=0.45\textwidth]{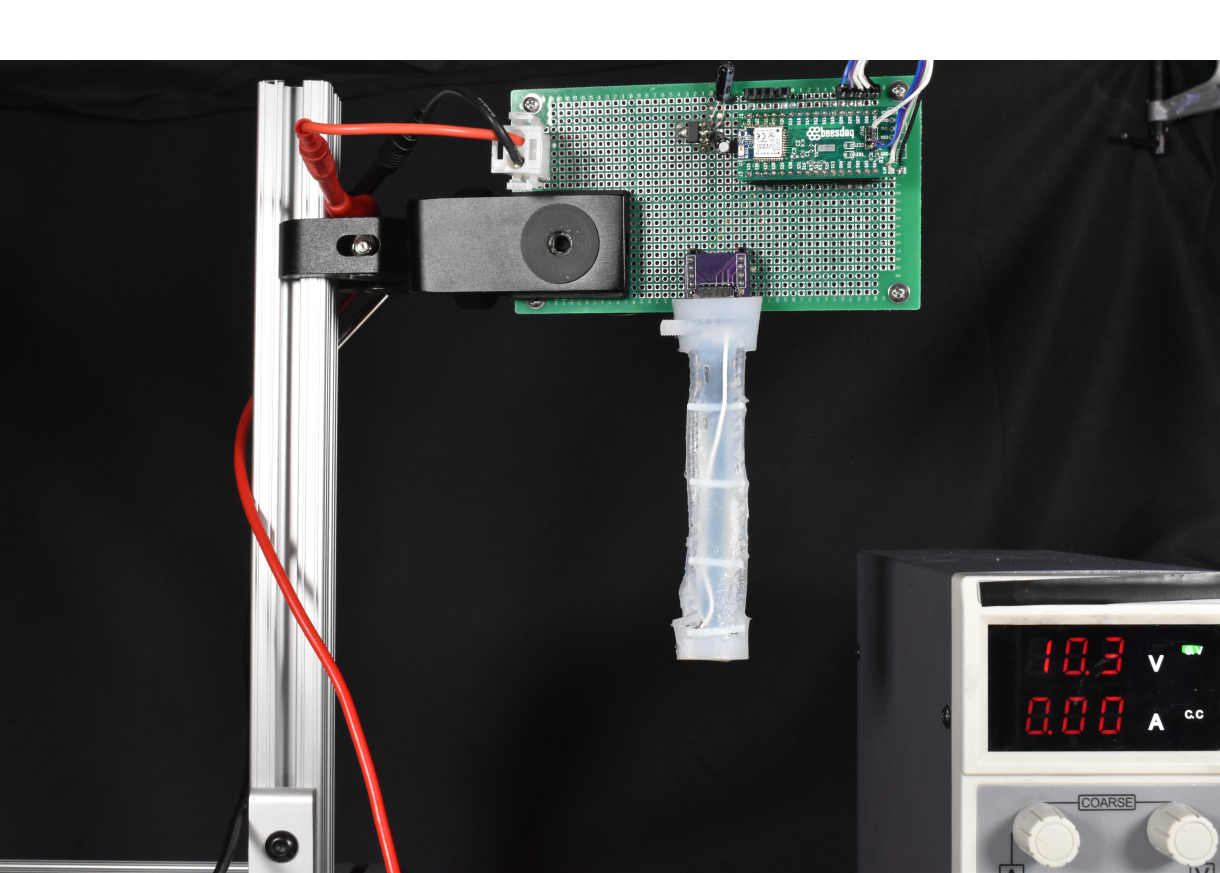}
\caption{Test setup for controller performance evaluation.}
\label{fig:setup}
\end{figure}

\begin{figure*}[tph]
\centering
\includegraphics[width=0.95\textwidth]{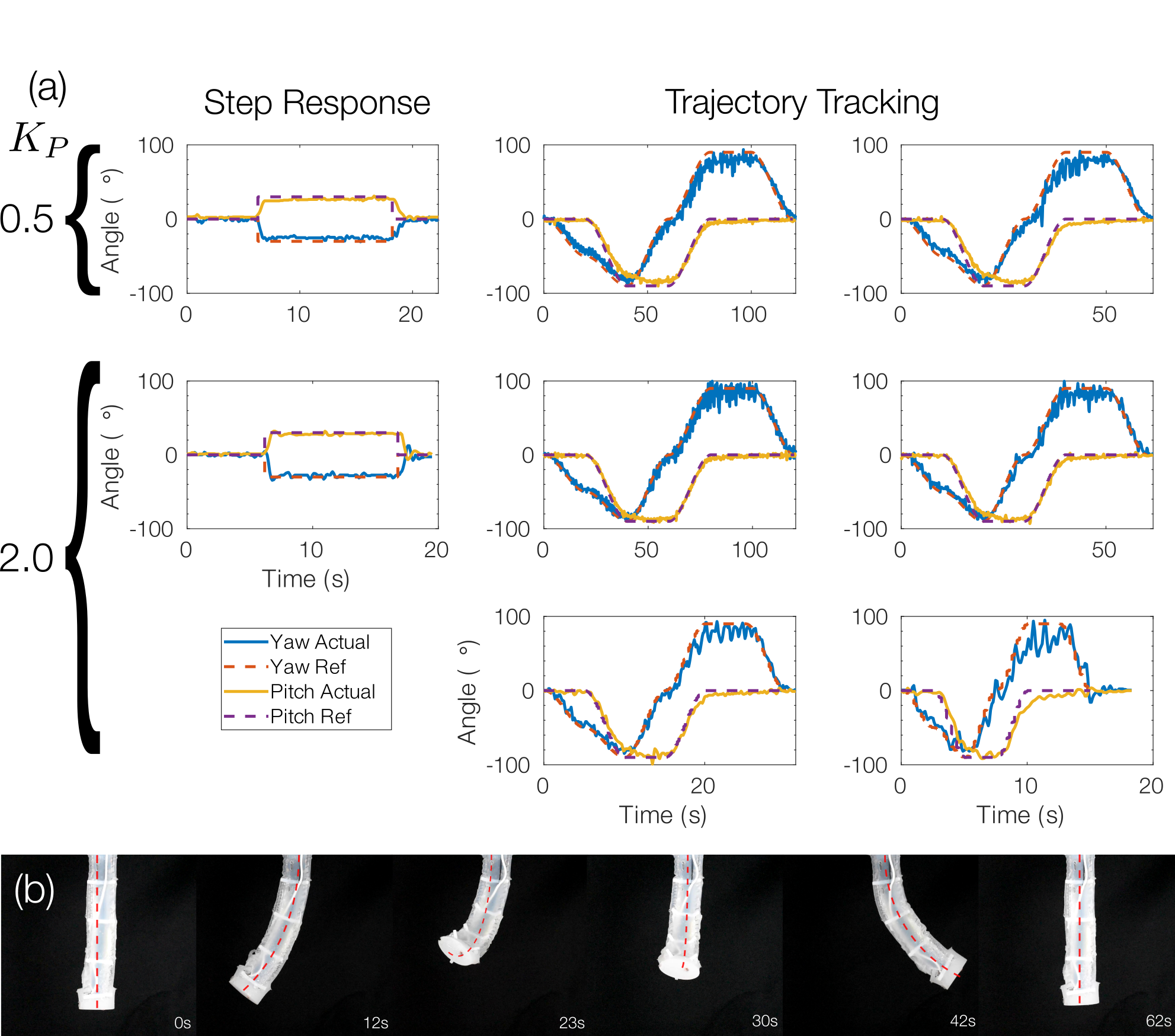}
\caption{Control results. (a) Step responses to a reference of 30$^\circ$ and trajectory tracking for gains $K_P = 2.0$ and $K_P = 0.5$. In every case, the shape of the trajectory is identical but the time over which it is executed is varied. (b) Snapshots of the manipulator during a one minute trajectory trial. Dashed red lines indicate the current desired bending shape.}
    \label{fig:trajectories}
\end{figure*}

\section{Controller Performance on Hardware}\label{results}
In this section, we present demonstrations of the controller from the previous sections on the SMA-actuated soft robot manipulator. We choose the gains discussed in the Stability Proof section, namely, $K_I = 1.5$ and $K_P = 2.0$ or $K_P = 0.5$. 

The test setup is presented in Fig. \ref{fig:setup} and described as follows. The manipulator is attached to a custom perfboard containing the microcontroller and power management electronics. This assembly is attached to a simple 80/20 assembly with the manipulator hanging down. A remote computer sends commands over bluetooth in the form of desired bend angles. The controller itself is implemented in its state space form on the microcontroller. The actuators are powered by a power supply set to 10.3V and maximum current output is limited to 2.5A. 

First, we show simple step response plots in Fig. \ref{fig:trajectories}a for a step of 30 degrees in both pitch and yaw for each of the two proportional gains. Of particular interest is the steady state error for both cases. For $K_P = 2.0$, the average yaw error is 2.19$^\circ$ and average pitch error is 1.94$^\circ$ while for $K_P = 0.5$, the average yaw error is 4.65$^\circ$ and average pitch error is 3.74$^\circ$.

After testing the step responses, we characterized the trajectory tracking for each of the two cases for various speeds. Fig. \ref{fig:trajectories}a also shows these results. The shape of the trajectory is the same for each case, but we test it over successively shorter durations. Testing with a gain of $K_P = 0.5$ is performed over two minutes and one minute. Testing is then done with a gain of $K_P = 2.0$ over the same as well as over 30 seconds and 15 seconds. It can be observed in the Figure as well as Table 1 that the higher gain achieves better performance, and that performance degrades as the trial speed increases. Finally, Fig. \ref{fig:trajectories}b shows images at several time steps of one of the trajectory trials.



\begin{table}[t]
\centering
\begin{tabular}{||c c c c||} 
 \hline
 $K_P$ & Duration (s) & Yaw Error ($^\circ$) & Pitch Error ($^\circ$) \\ [0.8ex] 
 \hline\hline
 0.5 & 120 & 8.44 & 4.84 \\ 
 \hline
 0.5 & 60 & 9.37 & 5.52 \\
 \hline
 2.0 & 120 & 5.18 & 2.93 \\
 \hline
 2.0 & 60 & 6.43 & 3.84 \\
 \hline
 2.0 & 30 & 7.35 & 4.69 \\
 \hline
 2.0 & 15 & 12.02 & 6.68 \\
 \hline
\end{tabular}
\caption{Error during trajectory trials for various gains and speeds.}
\end{table}

\section{Discussion}\label{discussion}

Our results exhibit low-error tracking performance, consistently tracking signals within 5$^\circ$ to 10$^\circ$ error at most, even for fast commands and large deformations.
This performance is close to the uncertainty of our bending sensor itself.
In addition, this performance is in line with the state of the art in the literature for both SMA-driven flexible manipulators as well as soft manipulators in general \cite{wang_parameter_2019,jarrett_robust_2017,alqumsan_robust_2019,doroudchi_tracking_2021}. 
The work presented here provides a new solution for feedback control of the unique combination of a soft robot limb, moving along multiple axes, powered by shape-memory actuators.

The controller used here enables this three-dimensional state-feedback bending control with a simple system model, avoiding the electro-thermo-mechanical dynamics of the SMA \cite{brinson_one-dimensional_1993}, hyperelasticity of the silicone rubber structure, and large deformations of the beam away from the modeled equilibrium. 
Our approach demonstrates that simple LTI models, combined with robust control, can achieve similar results as complicated models with simple control.

Three particular areas of improvement are possible from these results.
First, more optimal controllers may be possible with our current framework: we would like to move from robust stability to robust \textit{performance}.
Our stability analysis from Section \ref{proof} resulted in reasonable and well-performing controllers, but they are inherently conservative due to our analysis technique \cite{campo_robust_1990}).
Future work will use a less conservative stability proof that incorporates more knowledge about the structure of actuator saturation \cite{kothare_multiplier_1999}. 

Second, improvements to the model within our framework may address some of the errors observed in tracking.
From Fig. \ref{fig:trajectories}, the oscillations in control are larger when the manipulator is further from its equilibrium. 
Our Euler Bernoulli beam model assumes small deflections, so future work will examine if a large deflection beam model (e.g. Timoshenko \cite{timoshenko_lxvi_1921}) will provide a more accurate plant for larger regions of the state space. 
In addition, future work will focus on explicitly modeling and controlling the velocity dynamics.

Finally, as the core of our controller is a PI block, the current implementation artificially induces stiffness to achieve low-error tracking \cite{della_santina_controlling_2017}.
As our compact soft robots are intended for nondestructive interactions with their environment, future work will examine methods to reduce this artificial stiffness.
Fortunately, the PI controller is the least important part of our infrastructure; we can easily include any SISO control element in its place.
Therefore, future work will include more substantial feedforward elements to balance out the stiffness induced by the intergrator, as well as internal actuator state sensing for anti-windup in the presence of disturbances. 


\section{Conclusion}

This article contributes a robust control approach to multi-axis soft robotic limbs powered by shape memory alloy (SMA) actuators.
The results demonstrate, for the first time, state feedback in hardware for such a robotic manipulator.
The controller introduces several feedback elements - including singular value decomposition control and MIMO anti-windup - for use in soft robotics.
This approach provides an out-of-the-box option for control of a wide variety of soft manipulators with novel actuators, without the need for machine learning or system identification particular to a specific robot.
In doing so, these results provide a pathway for state feedback of more complicated SMA-powered soft robots, including multiple links in a serial chain, with future application to untethered legged soft robots.






\bibliographystyle{ieeetr}
\bibliography{Robosoft_RAL_2022}

\end{document}